\pdfoutput=1

\documentclass[11pt]{article}

\usepackage[preprint]{acl}

\usepackage{times}
\usepackage{latexsym}
\usepackage{amssymb}  
\usepackage{hyperref}
\usepackage{graphicx}
\usepackage{afterpage}

\usepackage{amsmath}
\usepackage{microtype}
\usepackage{xurl}
\usepackage[T1]{fontenc}

\usepackage[utf8]{inputenc}

\usepackage{microtype}

\usepackage{setspace}
\usepackage{inconsolata}

\usepackage{graphicx}
\usepackage{afterpage}
\usepackage[switch]{lineno} 
\usepackage{enumitem}

\title{\includegraphics[height=1.25em]{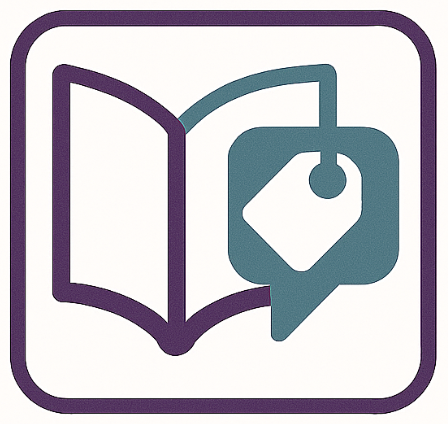} EduCoder: \\ An Open-Source Annotation System for Education Transcript Data}

\author{
  \textbf{Saad Ashraf\textsuperscript{1}},
  \textbf{James Malamut\textsuperscript{1}},
  \textbf{Vishal Kumar\textsuperscript{1}},
  \textbf{Guanzhong Pan\textsuperscript{2}},
  \textbf{Hyunji Nam\textsuperscript{1}},
  \textbf{Mei Tan\textsuperscript{1}},\\
  \textbf{Lucía Langlois\textsuperscript{1}},
  \textbf{Liliana Deonizio\textsuperscript{1}},
  \textbf{Helen Higgins\textsuperscript{1}},
  \textbf{Dorottya Demszky\textsuperscript{1}}
\\
\\
\textsuperscript{1}Stanford University,
  \textsuperscript{2}Carnegie Mellon University
}

\begin{document}
\maketitle

\begin{abstract}
We present \includegraphics[height=1.25em]{Figures/pure_icon.png} EduCoder, an open-source\footnote{\href{https://github.com/EduNLP/EduCoder}{https://github.com/EduNLP/EduCoder}} web platform designed for annotating classroom conversation transcripts. Existing annotation tools do not support the team-based workflows or access to instructional context that education discourse research requires. EduCoder addresses these gaps by combining transcript text, synchronized video, and instructional materials within a single workspace. The platform supports scoping annotation to specific portions of a lesson, coordinating work across annotation teams, and optionally integrating LLM-generated annotations with structured human--LLM comparison. EduCoder is freely accessible at \href{https://edu-coder.com}{edu-coder.com} and demonstrated in an accompanying video.\footnote{\href{https://www.youtube.com/watch?v=L4MzvmylgW0}{https://www.youtube.com/watch?v=L4MzvmylgW0}}

\end{abstract}

\afterpage{%
\begin{figure*}[t]
\centering
\includegraphics[width=\textwidth]{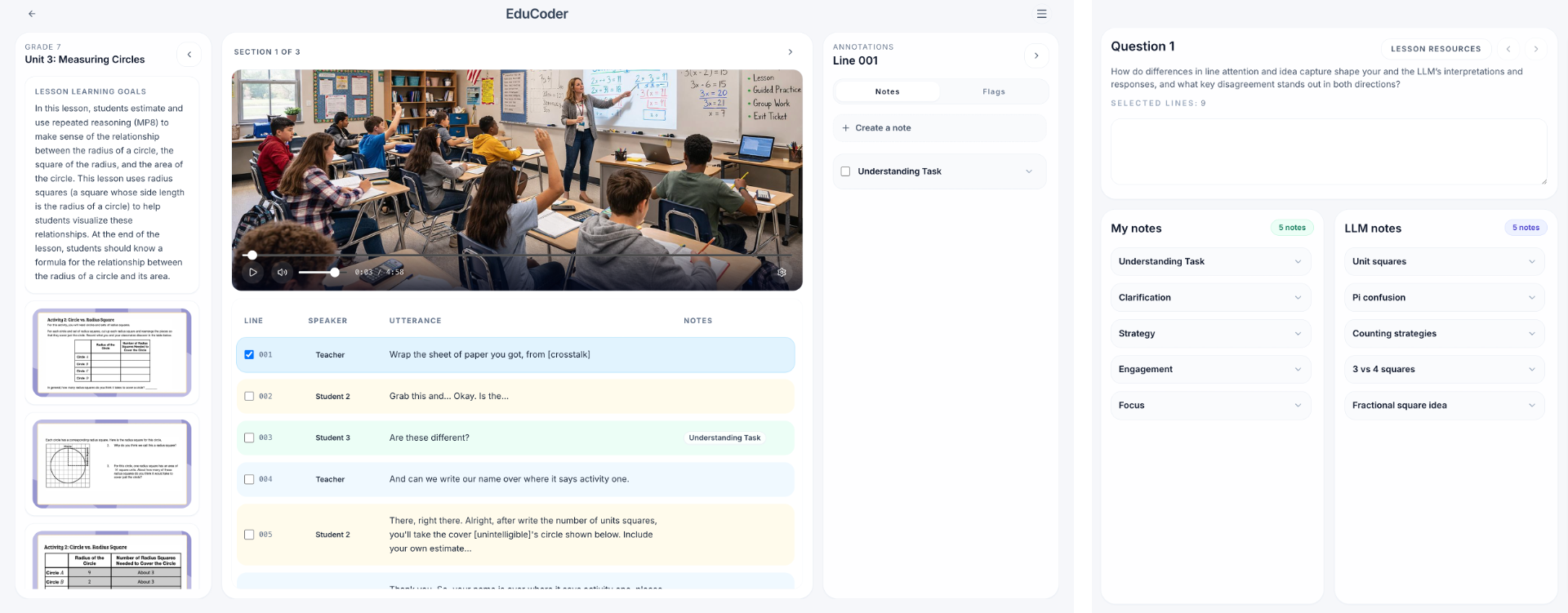}
\caption{EduCoder interface showing the annotation workspace (left) and the post-annotation comparison screen (right)}
\label{fig:interface}
\end{figure*}}

\section{Introduction}

Teacher--student and student--student dialogue play a critical role in learning, offering insight into instruction quality, student reasoning, and understanding \citep{Mercer2012, Michaels2008}. In recent years, the release of an increasing number of conversational datasets reflects growing interest in analyzing such dialogues \citep{ stasaski2020cima, suresh2022talkmoves, demszky2023ncte, wang2024bridging, macina2023mathdial}. Since manual analysis is labor-intensive and difficult to scale, there has been growing effort to develop natural language processing (NLP) based measures of educationally relevant constructs in transcript data \citep{d2012language,jensen2021deep,hunkins2022beautiful,demszky2021measuring,alic2022computationally,suresh2018using,zhao2024towards}. However, progress in NLP model development for analyzing teachers' and students' talk moves hinges on thoughtful human annotation, which is critical for model validation and training.

\paragraph{Challenges \& Motivation.}
Despite the availability of general-purpose annotation platforms, three persistent challenges remain in educational transcript annotation.

First, annotation workflows lack access to instructional context. Classroom annotation rarely involves the transcript alone; coders must consult lesson goals, instructional materials, and video evidence to make well-founded judgments. When these resources reside in separate tools, annotators must repeatedly switch contexts, which slows work and introduces inconsistency.

Second, classroom lessons are not uniformly relevant for annotation. Research teams typically focus on specific instructional episodes---a warm-up activity, a group-work segment, a whole-class discussion---that are defined by time ranges. Many workflows resort to manual video trimming or ad hoc timestamp navigation to constrain attention to annotation-relevant portions, a process that is time-consuming, error-prone, and difficult to reproduce.

Third, there is growing interest in leveraging large language models (LLMs) as annotation aids, yet existing tools offer limited support for structured human--LLM integration. LLMs have shown promising performance on educational content classification and other labeling tasks, sometimes approaching inter-rater reliability comparable to human coders \citep{bojic2025evaluating, long2024evaluating}. However, their reliability remains less understood when applied to more complex or socially nuanced constructs. Recent work has proposed treating LLM outputs as additional ``raters'' to help quantify and control measurement error \citep{broska2024mixed,calderon2025alternative}. However, teams currently lack mechanisms to control when and how model-generated references are revealed to annotators, which determines whether LLM outputs serve as aids or introduce bias. They also lack ways to systematically compare human judgments against model interpretations in a way that is aligned with research design goals.

\paragraph{Our system.} To address these challenges, we introduce EduCoder, an open-source web platform designed specifically for classroom transcript annotation. EduCoder provides:
\begin{itemize}[nosep,leftmargin=*]
    \item An \textbf{integrated annotation workspace} combining transcripts, synchronized video, and instructional context;
    \item \textbf{Flexible segmentation} that lets researchers defines which parts of a transcript to annotate directly from transcript metadata;
    \item \textbf{Team-based workflows} with configurable roles (e.g. admin, annotator) to support the coordination needs of education research teams.
    \item \textbf{LLM integration tools} with controls over when model-generated labels are shown to annotators, and how they are compared to human judgments.
\end{itemize}
\paragraph{} 
EduCoder is open-source software released under the MIT License and is freely accessible at \url{edu-coder.com}. Users can create a workspace directly on the platform and begin annotating without any local installation, lowering the barrier to entry for classroom discourse research teams.

\begin{table*}[ht]
\centering
\resizebox{\textwidth}{!}{%
\begin{tabular}{|l|c|c|c|c|c|c|c|c|c|c|c|}
\hline
\textbf{Feature} & \textbf{\texttt{EduCoder}} & \textbf{INCEpTION} & \textbf{Prodigy} & \textbf{Doccano} & \textbf{Label Studio} & \textbf{LightTag} & \textbf{POTATO} & \textbf{AWOCATo} & \textbf{EASE} & \textbf{AutoDive+} \\
\hline
Integrated Instructional Context & \checkmark &  &  &  &  &  &  &  & \checkmark & \\
Utterance-Level Annotation & \checkmark & \checkmark & -- & -- & -- & -- & -- & \checkmark & \checkmark & \checkmark \\
Segment-Aware Navigation & \checkmark &  &  &  &  &  &  &  &  &  \\
Transcript--video Synchronization & \checkmark &  & -- &  &  &  & -- &  &  &  \\
Admin-controlled LLM reference notes & \checkmark & -- & -- &  & -- & -- & -- &  & -- & -- \\
Structured Human--LLM Comparison & \checkmark &  &  &  &  &  &  &  &  &  \\
Role-based workflow tracking & \checkmark & \checkmark & -- & -- & -- & -- & -- & \checkmark &  &  \\
Open source & \checkmark & \checkmark &  & \checkmark & \checkmark &  & \checkmark & \checkmark & \checkmark & \checkmark \\
\hline
\end{tabular}%
}
\caption{Comparison of \texttt{EduCoder} with widely used annotation tools across key capabilities relevant to educational dialogue annotation. 
A checkmark (\checkmark) indicates the tool fully supports the feature. 
A dash (--) indicates the tool provides the functionality at a basic or partial level but does not meet the specific requirements. 
A blank cell indicates the feature is not supported.}
\label{tab:annotation_comparison}
\end{table*}

\section{Related Works}
\paragraph{General-purpose annotation tools. }
The NLP community has benefited greatly from general-purpose annotation toolkits like INCEpTION \citep{Klie2018}, Prodigy \citep{prodigy}, Label Studio \citep{tkachenko2020labelstudio}, and Doccano \citep{nakayama2018doccano}, which standardize data labeling tasks such as named entity recognition and text classification. The newer versions of Label Studio and Prodigy also feature LLM integrations for machine coding. However, these tools are designed primarily for labeling independent spans within documents, rather than for the structured, open-ended coding of conversational transcripts common in education dialogue research.

\paragraph{Annotation tools with advanced features. }
Recent tools like  AWOCATo \citep{daudert2020web}, LightTag \citep{perry2021lighttag}, POTATO \citep{pei2022potato}, EASE \citep{deng2023ease} and AutoDive+ \citep{wang2025autodive+} offer more advanced annotation features. AWOCATo provides customizability through JSON configurations, including free-text and categorical labeling. LightTag features annotator productivity enhancements such as batching and pre-annotation alongside inter-annotator agreement metrics. POTATO offers extensive customization through task templates, active learning, and quality control mechanisms. EASE integrates multi-task active learning with direct LLM integration. AutoDive+ supports multimodal annotation, active learning, and a community-driven model repository. Despite these strengths, none of these tools offer extensive contextual referencing needed to preserve the sequential nature of transcripts and account for additional contextual information.

\paragraph{Human--LLM collaboration in annotation}
The integration of human judgment with AI systems has gained significant attention in NLP, particularly in reframing human-AI collaboration for explanatory tasks \citep{wiegreffe-etal-2022-reframing}. Recent work has explored how LLMs parse and interpret text compared to human annotators \citep{hong-etal-2024-large}, highlighting the importance of understanding differences in human versus AI text processing patterns \citep{munoz-ortiz2023contrasting}. This body of research informs our approach to incorporating LLM-generated annotations as a reference layer while maintaining human expertise as the primary decision-making authority. 

\afterpage{%
\begin{figure*}[t]
\centering
\includegraphics[width=\textwidth]{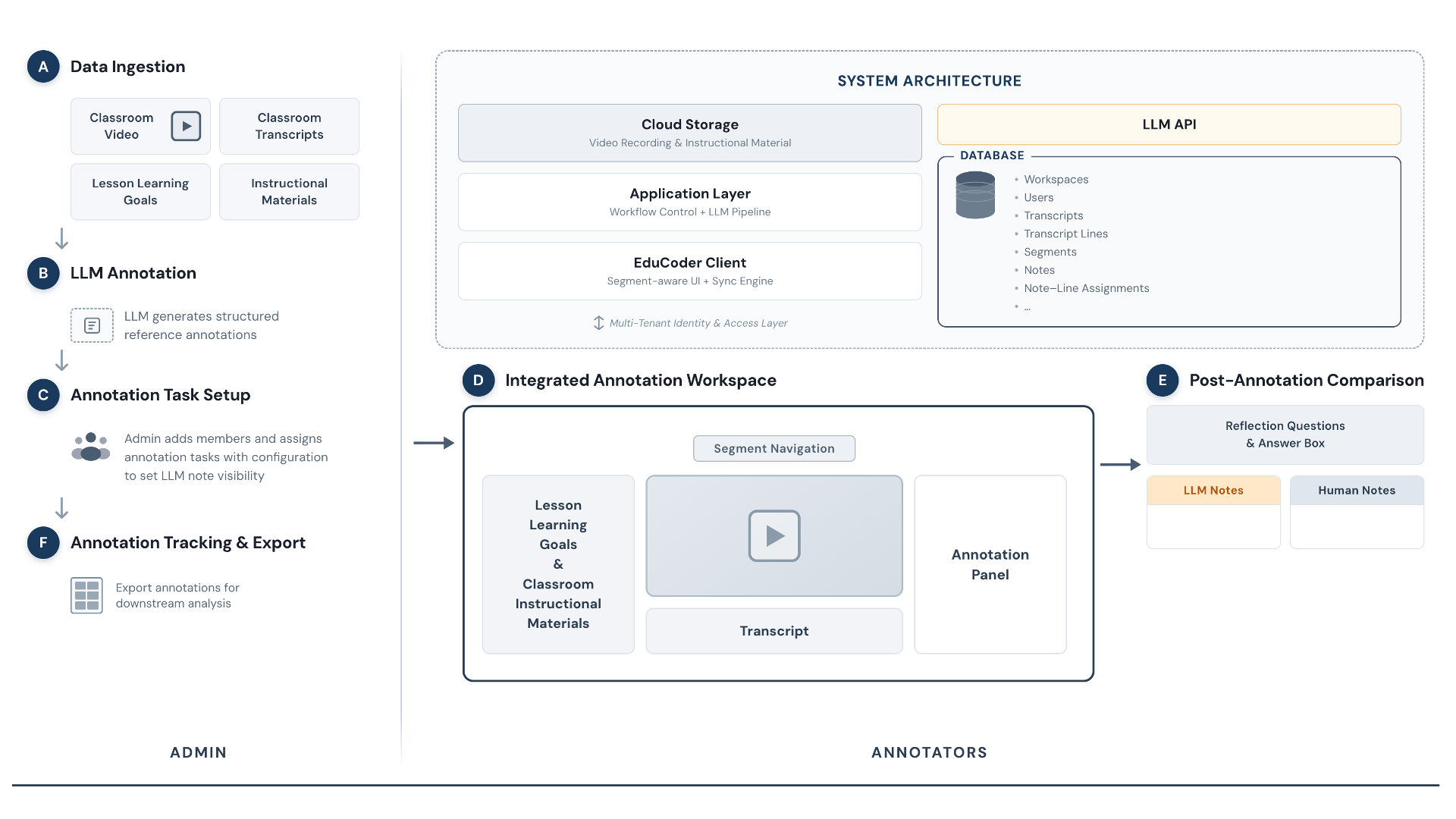}
\caption{EduCoder system architecture and annotation workflow. The left column shows admin-driven steps and the right column shows the system architecture and annotator-facing views}
\label{fig:pipeline}
\end{figure*}}

\afterpage{%
\begin{figure*}[t]
\centering
\includegraphics[width=\textwidth]{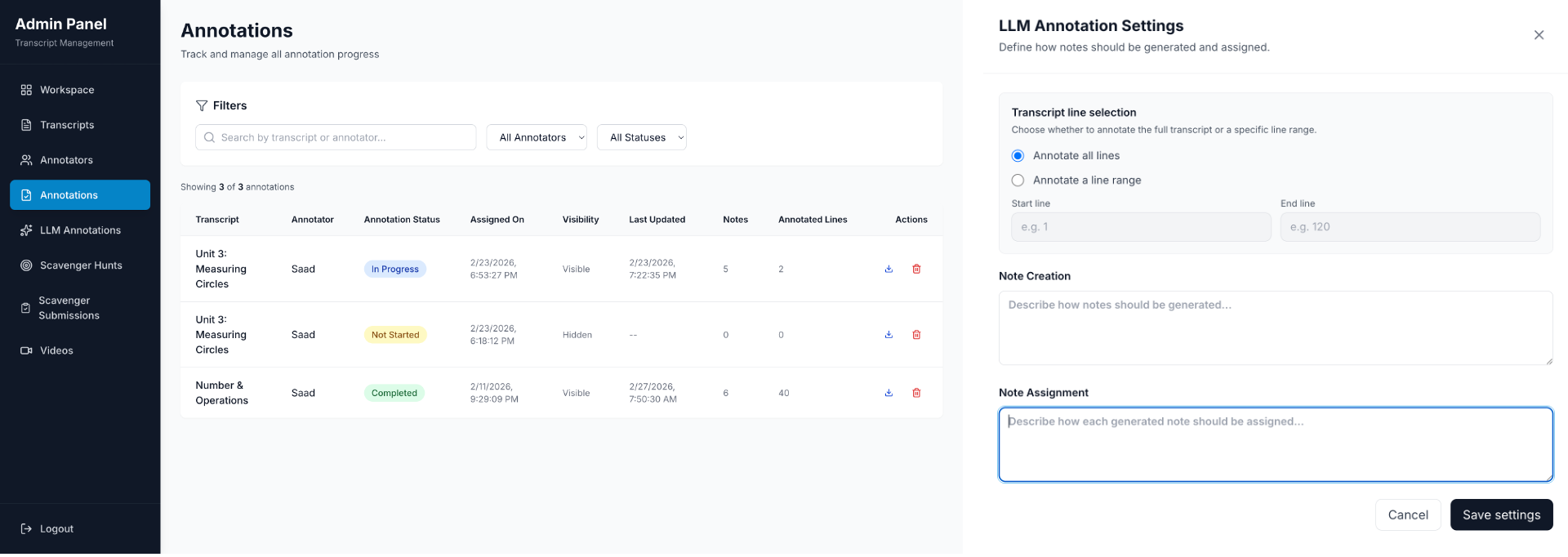}
\caption{Admin interface for managing annotation progress (left) \& LLM annotation configuration panel (right).}
\label{fig:3}
\end{figure*}}

\section{System Overview}
EduCoder supports an end-to-end classroom transcript annotation workflow, illustrated in Figure~\ref{fig:3}. An administrator can upload transcripts alongside supporting materials such as classroom video recordings, images of student work, and learning goals. The admin may optionally generate LLM reference annotations and then configure annotation tasks, including role assignments for annotators and LLM note visibility settings. Annotators work within an integrated workspace that combines the transcript, synchronized video, and instructional context, creating notes and attaching them to specific transcript lines within segment-scoped views. After annotation, annotators can engage in a structured comparison activity that surfaces agreement and divergence between human and LLM interpretations. Throughout this process, admins monitor progress and export completed annotations for downstream analysis.

\subsection{Architecture}

EduCoder is implemented as a multi-tenant full-stack web system. The platform uses a Next.js App Router application for both the admin and annotator interfaces, with server-side API routes handling business logic. This design reduces integration overhead and allows role-specific workflows (admin, annotator, and system/LLM roles) to be enforced at the API layer. Authentication supports both email-link sign-in and workspace credential login to separate configuration, annotation, and system-generated outputs.

The backend is organized as a service layer over PostgreSQL using Prisma ORM. Core relational entities include workspaces, users, transcripts, transcript lines, transcript segments, annotations, notes, and note-to-line assignments. Workspace IDs are used throughout to enforce tenant isolation and scope reads and writes to the current team, supporting multi-team usage on shared infrastructure while preserving data separation. Large artifacts are stored in Google Cloud Storage while metadata resides in PostgreSQL; for videos, the system uses signed upload and playback URLs, keeping application servers stateless.

\paragraph{Workspace and roles.} EduCoder organizes work around two roles---admins and annotators---to support team-based annotation workflows. When a new user signs in, the system initializes a workspace and assigns them as admin. Admins invite members, assign roles, manage transcript assignments, and control feature visibility such as whether LLM-generated notes are shown to annotators. Annotators complete assigned transcripts by producing notes and marking their work as complete.

\paragraph{Transcript and artifact uploads.} EduCoder supports a wide variety of input formats to minimize preparation effort for users. Transcripts can be uploaded as \texttt{.csv}, \texttt{.xls}, or \texttt{.xlsx} files containing dialogue text, speaker information, and optional timing or segmentation metadata. The system also accepts video artifacts in a broad range of formats, allowing researchers to work with existing classroom recordings without conversion. During upload, EduCoder performs lightweight validation and provides clear feedback if any issues would prevent reliable annotation.

\paragraph{Data model and transcript--video synchronization.} Internally, EduCoder represents a transcript as an ordered collection of lines, where each line stores speaker attribution, utterance text, and optional timing metadata. Annotation is represented as notes authored by annotators, alongside a many-to-many note-to-line mapping that records which lines each note addresses. When time cues are provided (as seconds or timecodes), EduCoder automatically normalizes them to enable synchronization with the associated video: selecting a line updates the playback position, and playback position can be used to orient annotators within the transcript. When timing metadata is absent, EduCoder falls back to transcript-only navigation while preserving the same note and mapping structure.

\paragraph{Segment-aware navigation.} A central design feature of EduCoder is its segment-aware transcript mechanism, which uses transcript metadata---rather than manual video trimming---to define annotation-relevant spans. If a \textit{Segment} column is provided, EduCoder constructs an ordered list of segments and associates each transcript line with a segment label. Segment timing is inferred from the cues of lines contained in the segment (e.g., the earliest in-cue and latest out-cue among member lines), producing a segment-level span that can be navigated directly. This enables annotators to work within instructional phases such as warm-up, group work, or whole-class discussion.

When segment metadata is not present, EduCoder operates in an unsegmented mode with a full transcript view. When cues are partially missing, the system degrades gracefully by maintaining segment structure for navigation while disabling time-based video jumps for lines without reliable cues. By deriving segments directly from the transcript, EduCoder eliminates the need for a separate video clipping pipeline.

\subsection{Annotator Interface}

EduCoder's annotator workspace is organized into a three-column layout designed to keep instructional grounding and evidence accessible throughout the labeling process. The left panel presents lesson learning goals and instructional materials (e.g., prompts, worksheets). The center panel provides a synchronized transcript and video view with segment navigation, enabling annotators to move efficiently between activities and evidence. The right panel supports annotation actions: creating notes, selecting one or more transcript lines, attaching notes to lines, and adding flags.

The interface emphasizes utterance-level operations with clear speaker differentiation and lightweight interaction primitives that support rapid evidence gathering. A menu provides optional features such as keyword search, speaker filtering, and show/hide controls for specific parts of the transcript table. The video player supports viewing note placements as colored bars directly on the video timeline, and all annotations are auto-saved to the database.

\paragraph{Admin monitoring and export.} EduCoder provides admins with progress monitoring dashboards oriented toward multi-annotator workflows. At the transcript level, admins can track assignment state (e.g., not started, in progress, completed) and review summary indicators such as note counts and annotated-line coverage. Admins can also inspect the distribution of work across annotators to identify bottlenecks or transcripts requiring follow-up. For downstream analysis, EduCoder supports exporting annotations in spreadsheet format.

\paragraph{LLM-assisted annotation and structured reflection.} EduCoder provides support for integrating LLM-generated annotations into human workflows. This integration comprises two modules: an LLM reference note pipeline for generating and selectively revealing model annotations, and a Scavenger Hunt Comparison Activity for structured post-annotation reflection. Both modules are described below.

\paragraph{LLM reference note pipeline.} EduCoder allows administrators to generate LLM reference annotations for any transcript in the workspace. Notes can be generated for an entire transcript or for selected lines. EduCoder provides a default annotation prompt, which admins can customize.

LLM note generation proceeds in two stages. First, the model produces candidate notes given the transcript text and any available context (e.g., lesson goals, materials). Second, each note is assigned to one or more transcript lines. This assignment is done segment by segment so the model stays grounded in nearby context rather than drawing spurious connections across distant parts of the transcript. We experimented with a single-step approach but found that separating note creation from line assignment produced better notes with more accurate transcript grounding.

Because LLM-generated notes could bias annotators toward the model's suggestions, EduCoder provides controls over who sees them and when. Administrators can configure per-annotator visibility settings so that LLM notes are available during annotation, only after annotation is marked complete, or not at all. This enables designs such as showing LLM references to a calibration subset of annotators while withholding them from others, or using them exclusively in post-annotation review.

\paragraph{Post-annotation comparison activity: Scavenger Hunt.}

After human annotation is complete, EduCoder supports a structured comparison activity called the \textit{Scavenger Hunt}, in which annotators review their notes alongside LLM-generated notes. This activity is motivated by three complementary goals. First, \textbf{meta-reflection}: by explicitly comparing their annotations against an LLM's interpretation, annotators are prompted to articulate the reasoning behind their coding decisions, surface implicit assumptions, and reconsider ambiguous cases. Second, \textbf{research on human–LLM alignment}: teams can analyze where model and human annotations systematically agree or diverge, generating actionable feedback for improving automated annotation systems and human coding protocols. Third, \textbf{structured reflective artifacts}: the collected comparisons can support professional development workflows — for example, teacher educators can use the activity to guide pre-service teachers through close analysis of classroom transcripts, with LLM annotations serving as discussion scaffolds.

The activity presents annotators with a sequence of guided comparison questions designed to surface agreement, divergence, and reasoning differences between human and model interpretations. For each comparison prompt, annotators provide free-text responses and may link relevant transcript lines as supporting evidence. Responses are auto-saved, and the system tracks completion status per annotator and per transcript. Admins can export submissions as spreadsheets for downstream analysis.

\section{Evaluation and Case Study}
To evaluate EduCoder’s usability and effectiveness,
we conducted a focus group at a university summit
for educators interested in the application of language technology in education. The study involved
16 K-12 mathematics teachers and 2 university based math education professors. Participants completed a 45-minute training session followed by
a 60-minute collaborative annotation task using
common classroom dialogue transcripts. The codebook included both categorical features (e.g., identifying student discourse roles) and free-response fields capturing annotators' observations of students' mathematical reasoning. The annotation workflow incorporated calibration exercises on one to three reference transcripts to establish baseline agreement before independent coding.

Of the 18 participants, 12 completed a post-session survey capturing adoption likelihood, preferred usage contexts, and perceived barriers.

\paragraph{Adoption Likelihood. }

All 12 respondents indicated willingness to adopt AI-assisted classroom discourse analysis: 8 (66.7\%) rated adoption as \emph{extremely likely} and 4 (33.3\%) as \emph{somewhat likely}, with no negative ratings. Teachers who rated adoption as \emph{extremely likely} cited the ability to access student conversations they would otherwise miss during instruction and support for analyzing small-group mathematical thinking.

\paragraph{Preferred Usage Contexts. }

Preferences were distributed across private self-reflection (33.3\%), collaborative PLC or grade-level team settings (25.0\%), one-on-one instructional coaching (25.0\%), and multi-context use (16.7\%). This distribution suggests no single deployment mode dominates and reinforces EduCoder's design decision to support flexible usage configurations.

\paragraph{Barriers.}

Privacy and data concerns were the most frequently cited barrier (41.7\%), followed by concerns about AI accuracy in identifying nuanced mathematical talk moves (25.0\%), cost and infrastructure limitations (16.7\%), and time constraints (16.7\%). These findings validate EduCoder's architecture choices around data security and human-in-the-loop design.

 Our results suggest that teachers see clear value in AI-assisted discourse analysis but want it on their own terms — across varied professional contexts and with safeguards around privacy and interpretive accuracy. The strongest theme across responses was access: teachers described the tool as a way to surface student thinking they would otherwise miss during live instruction. Participants also described the annotation process itself as a catalyst for reflection. As one teacher noted, "It makes me really reflect on who is speaking and how different dynamics of student groups (gender, cultural, language, etc.) can affect their collaborative conversations." The spread of preferred contexts, from private reflection to coaching to team-based PLC work, supports EduCoder's design as a flexible platform that can serve both research annotation and practitioner professional development.

\section{Conclusion}

We introduce EduCoder, an open-source web platform for annotating classroom conversation transcripts. EduCoder provides a unified workspace for utterance-level annotation and configurable human--LLM workflows that allow teams to incorporate model-generated references while preserving human interpretive authority. By providing an accessible, collaborative environment, EduCoder aims to support both research teams and practitioners in analyzing classroom discourse at scale.

\paragraph{Limitations and future work.} Our work has limitations in both the scope of our evaluation and the tool's current capabilities. Our evaluation involved 18 participants in a single session; longer-term deployment studies are needed to understand how annotation workflows evolve over time and scale. On the technical side, transcript uploading requires predefined column mappings; future versions will support schema-free parsing to accommodate a wider variety of formats. Furthermore, the system does not yet include built-in personally identifiable information (PII), redaction. Finally, the LLM annotation pipeline operates solely on transcript text; incorporating video context into model-generated annotations is a promising direction for improving alignment with human interpretations that draw on multimodal evidence.

\section{Ethical Considerations}

EduCoder is designed to support annotation of classroom conversation transcripts, which may involve recordings of teachers and students, including minors. We outline the key ethical dimensions of our work below.

\paragraph{Privacy and data protection.}
Workspace administrators are responsible for ensuring that they have appropriate permissions and institutional approval (e.g., IRB clearance) to upload classroom data to EduCoder. While data is stored with encryption and workspace-level isolation ensures that annotation data is not shared across teams, administrators should review our privacy policy and verify that the platform's data handling practices are compatible with their institutional requirements before uploading any sensitive material. For teams with stricter data governance needs, EduCoder's source code is publicly available, allowing institutions to deploy and manage their own instances with full control over data storage and access.

\paragraph{Accuracy and fairness.}
EduCoder features LLM-generated reference annotations, but we cannot guarantee their accuracy across all settings. Administrators and users are responsible for evaluating whether the LLM outputs are reliable for their specific data. This is especially important when working with transcripts that represent underrepresented student or teacher populations, noisy or low-quality recordings, or classrooms conducted in languages other than English. In such cases, model performance may degrade in ways that could introduce systematic bias into annotation workflows. We strongly encourage teams to test LLM-generated annotations against human judgments on a representative sample of their data before incorporating model outputs into their workflow.

\bibliography{custom}

\appendix

\end{document}